\documentclass[aoas,preprint]{imsart}
\usepackage{fullpage}
\setattribute{journal}{name}{}

\usepackage{graphicx,verbatim}
\usepackage[round]{natbib}
\usepackage{url}
\usepackage{amsmath,amssymb,amsthm,amsfonts}
\usepackage{algorithm}
\usepackage{algpseudocode}
\usepackage{todonotes}
\usepackage{subfig}

\usepackage{tikz}
\usetikzlibrary{arrows}

\newcommand{\eat}[1]{ }

\def\ie{{\it i.e.},~}
\def\eg{{\it e.g.},~}

\def\E{\mathbb{E}}

\def\Unif{\mbox{Unif}}

\def\N{{\mathcal N}}
\newcommand{\x}{\mathbf{x}}

\newcommand{\nn}{\nonumber}
\newcommand{\be}{\begin{equation}}
\newcommand{\ee}{\end{equation}}
\newcommand{\bea}{\begin{eqnarray}}
\newcommand{\eea}{\end{eqnarray}}
\newcommand{\erf}{\text{erf}}

\newcommand{\given}{\,|\,}

\newcommand{\mcL}{\mathcal{L}}

\newcommand{\mcP}{\mathcal{P}}
\newcommand{\mcU}{\mathcal{U}}
\begin{document}

\begin{frontmatter}

\title{Accelerating MCMC via Parallel Predictive Prefetching}
\runtitle{Accelerating MCMC via Parallel Predictive Prefetching}

\begin{aug}
\author{\fnms{Elaine} \snm{Angelino}\ead[label=e1]{elaine@eecs.harvard.edu}},
\author{\fnms{Eddie} \snm{Kohler}\ead[label=e2]{kohler@seas.harvard.edu}},
\author{\fnms{Amos} \snm{Waterland}\ead[label=e3]{apw@seas.harvard.edu}},\\
\author{\fnms{Margo} \snm{Seltzer}\ead[label=e4]{margo@eecs.harvard.edu}}
\and
\author{\fnms{Ryan P.} \snm{Adams}\ead[label=e5]{rpa@seas.harvard.edu}}

\affiliation{Harvard University}
\runauthor{Angelino, Kohler, Waterland, Seltzer, and Adams}

\address{School of Engineering and Applied Sciences\\
Harvard University\\
Cambridge, MA\\
\printead{e1}\\
\phantom{E-mail:\ }\printead*{e2}\\
\phantom{E-mail:\ }\printead*{e3}\\
\phantom{E-mail:\ }\printead*{e4}\\
\phantom{E-mail:\ }\printead*{e5}}
\end{aug}

\begin{abstract}
We present a general framework for accelerating a large class of widely used Markov chain Monte Carlo (MCMC) algorithms.  Our approach exploits fast, iterative approximations to the target density to speculatively evaluate many potential future steps of the chain in parallel.
The approach can accelerate computation of the target distribution of a Bayesian inference problem, without compromising exactness, by exploiting subsets of data.
It takes advantage of whatever parallel resources are available, but produces results exactly equivalent to standard serial execution. In the initial burn-in phase of chain evaluation, it achieves speedup over serial evaluation that is close to linear in the number of available cores.
\end{abstract}

\end{frontmatter}

\section{Introduction}

Probabilistic modeling is one of the mainstays of modern machine learning.
Bayesian methods are especially appealing due to their ability to represent uncertainty in parameter estimates and latent variables.
Real-world problems are rarely amenable to exact inference, and so require approximate inference in the form of Monte Carlo estimates or variational approximations.
Unfortunately, approximate Bayesian inference can be challenging when modeling large data sets, as the target posterior density may become expensive to evaluate.  This challenge has motivated new methods for inferential computation that can take advantage of approximations to the target density,
most often by examining only a subset of the data,
or by exploiting closed form approximations
such as Taylor series~\citep{christen-fox-2005-approx},
or by fitting linear or Gaussian Process regressions~\citep{Conrad14}.
Stochastic variational inference techniques~\citep{Hoffman2013} achieve this by randomized approximations of gradients, while recent developments in Markov chain Monte Carlo (MCMC) have implemented efficient transition operators that
lead to approximate stationary distributions~\citep{Welling2011, Korattikara2014, Bardenet2014}.
Recent other work uses a lower bound on the local likelihood factor to simulate from the exact posterior distribution while evaluating only a subset of the data at each iteration~\citep{maclaurin-2014-firefly}.
We also focus on accelerating MCMC with approximations to the transition operator, but we arrive at a method in which the stationary distribution is \emph{exactly} the target posterior.

We attack the problem using parallelism. This is difficult, however:
MCMC algorithms such as Metropolis-Hastings (MH) are inherently serial and can be prohibitively slow to converge,
especially when the target function is high-dimensional and multi-modal. The embarrassingly parallel approach of running many independent chains does not decrease the mixing time for any single chain, and so tends not to reduce the time to achieve a useful estimator.  Sometimes the evaluation of the target function can be parallelized, or multiple chains in an ensemble method can be run in parallel, but these strategies are not available in general.

We instead attack the general problem by using speculative execution to parallelize MCMC algorithms.  This approach, sometimes called prefetching, has received some attention in the past decade, but does not seem to be widely recognized.  Consider the MH algorithm in Algorithm~\ref{mh}, in which each iteration consists of a proposal that is stochastically accepted or rejected~\citep{metropolis-1953}.  Given a source of randomness and an initial state, all possible future states of the chain can be thought of as the nodes of a binary tree (Figure~\ref{tree}).  Serial execution chooses a single path on the tree by executing nodes in sequence. Prefetching executes other nodes in the tree in parallel with the immediate transition (\ie at the root of the tree).

An effective prefetching implementation must overcome several challenges.
Some involve correctness. For example, for the results of prefetching to exactly equal those of a serial execution, care is required in the treatment of pseudo-randomness (\ie each node's source of randomness must produce the same results as it would in a serial execution); slapdash treatment risks introducing biases.
But the key challenge is performance. A na\"{i}ve scheduling scheme always requires
$\approx 2^s$ parallel cores to achieve a speedup of $s$. Previous improvements to this
speculative procedure~\citep{strid-2010-prefetching} can improve this speedup by
leveraging information about the average proposal acceptance rate.
In particular, if most proposals are rejected, a prefetching implementation can improve its speedup by prefetching more heavily along the reject path.
Although in practice the optimal acceptance rate is less than 0.5~\citep{Gelman96},
extremely small acceptance rates, which lead to good speedup, are accompanied by less effective mixing. If the optimal acceptance rate is set to something like 0.234, speedup is still at most logarithmic.

\begin{algorithm}[t]
\caption{Metropolis-Hastings}
\label{mh}
\begin{algorithmic}
\State \textbf{Input:} initial state $\theta_0$, number of iterations $T$, target $\pi(\theta)$, proposal $q(\theta' \given \theta)$
\State \textbf{Output:} Samples $\theta_1, \dots, \theta_T$\For {$t = 0, \dots, T-1$}
\State $\theta' \sim q(\theta' \given \theta_t)$
\State $u \sim \Unif(0, 1)$
\If {$\frac{\pi(\theta') q(\theta_t \given \theta')}{\pi(\theta_t) q(\theta' \given \theta_t)} > u$}
    \State $\theta_{t+1} = \theta'$
\Else
    \State $\theta_{t+1} = \theta_t$
\EndIf
\EndFor
\end{algorithmic}
\end{algorithm}

We evaluate a new scheduling approach that uses local information to improve speedup relative to other prefetching schemes.
We adaptively adjust speculation based not only on the local average proposal acceptance rate (which changes as evaluation progresses), but also on the actual random deviate used at each state.
Even better, we make use of any available fast approximations to the transition operator.
Though these approximations are not required, when they are available or learnable, we leverage them to make better scheduling decisions.

We present results using a series of increasingly expensive but more accurate approximations.
These decisions are further improved by modeling the error of these approximations, and thus the uncertainty of the scheduling decisions.
Performance depends critically on how we model the approximations, and a key insight is in our error model for this setting; much smaller error, and therefore more precise predictions, are obtained by modeling the error of the \emph{difference} between two proposal evaluations, rather than evaluating the errors of the proposals separately.
Motivated by large-scale Bayesian inference,
we present results using a series of increasingly expensive but more accurate approximations.
Our current implementation uses approximations that correspond to incremental evaluation of the target distribution, but our framework does not require this. 
As we show on inference problems using both real and synthetic data, our system takes advantage of parallelism to speed up the wall-clock time of serial Markov chain evaluation.
Unlike prior systems, we achieve near-linear speedup during burn-in on up to 64 cores.
As evaluation progresses, speedup eventually decreases to logarithmic in the number of cores; we show why this is hard to avoid.

\section{Parallel MCMC}

In Markov chain Monte Carlo, computational cost is most often determined by evaluation of the target density relative to mixing.  In Metropolis--Hastings, this cost is incurred when the target is evaluated to determine the acceptance ratio of a proposed move;
in slice sampling~\citep{Neal03}
an expensive target slows both bracket expansion and contraction. We focus on the increasingly common case where the target is expensive and the dominant computational cost.  This evaluation can sometimes be parallelized directly, \eg when the target function is a product of many individually expensive terms.  This sometimes arises in Bayesian inference if the target can be easily decomposed into one likelihood term for each data item.  Scalability (\ie practically achievable speedup) in this setting is limited by the communication and computational costs associated with aggregating the partial evaluations.  In general, the target function cannot be parallelized; we divide methods that accelerate MCMC via other sources of parallelism into two classes: ensemble sampling and prefetching.

\subsection{Ensemble Samplers}

Ensemble (or \emph{population}) methods for sampling run multiple chains and accelerate mixing by sharing information between the chains.  The individual chains can be simulated in parallel; any information sharing between chains requires communication. Examples include parallel tempering~\citep{swendsen-1986-tempering}, the emcee implementation~\citep{goodman-2012-emcee} of affine-invariant ensemble sampling~\citep{goodman-2010-affine}, and a parallel implementation of generalized elliptical slice sampling~\citep{nishihara-2012-gess}.

\subsection{Prefetching}

\begin{figure}[t!]
  \centering%
  \resizebox{\columnwidth}{!}{%
    \def\radius {5mm}
    \tikzstyle{state}=[circle, thick, minimum size=\radius, font=\footnotesize]
    \begin{tikzpicture}[->,>=stealth',level/.style={sibling distance = 5cm/#1, level distance = 1.5cm}]
      \draw [gray,-] (-6cm,-0.75cm) -- (6cm,-0.75cm);
      \draw [gray,-] (-6cm,-2.25cm) -- (6cm,-2.25cm);
      \draw [gray,-] (-6cm,-3.75cm) -- (6cm,-3.75cm);
      \node [state] {$\theta^t$}
      child{ node [state] {$\theta^{t+1}_0$}
        child{ node [state] {$\theta^{t+2}_{00}$}
          child{ node [state] {$\theta^{t+3}_{000}$}}
          child{ node [state] {$\theta^{t+3}_{001}$}}
        }
        child{ node [state] {$\theta^{t+2}_{01}$}
          child{ node [state] {$\theta^{t+3}_{010}$}}
          child{ node [state] {$\theta^{t+3}_{011}$}}
        }
      }
      child{ node [state] {$\theta^{t+1}_1$}
        child{ node [state] {$\theta^{t+2}_{10}$}
          child{ node [state] {$\theta^{t+3}_{100}$}}
          child{ node [state] {$\theta^{t+3}_{101}$}}
        }
        child{ node [state] {$\theta^{t+2}_{11}$}
          child{ node [state] {$\theta^{t+3}_{110}$}}
          child{ node [state] {$\theta^{t+3}_{111}$}}
        }
      }
      ;
      \node [state] at (5.5cm, 0) {$u^{t}$};
      \node [state] at (5.5cm, -1.5) {$u^{t+1}$};
      \node [state] at (5.5cm, -3) {$u^{t+2}$};
      \node [state] at (5.5cm, -4.5) {$u^{t+3}$};
    \end{tikzpicture}
  }
  \caption{Schematic of the prefetching state tree for Metropolis--Hastings.
  Each level of the tree represents an iteration, where branching to the right/left indicates that the proposal is accepted/rejected.
  The random variates (on right) are shared across the layer.}
  \label{tree}
\end{figure}
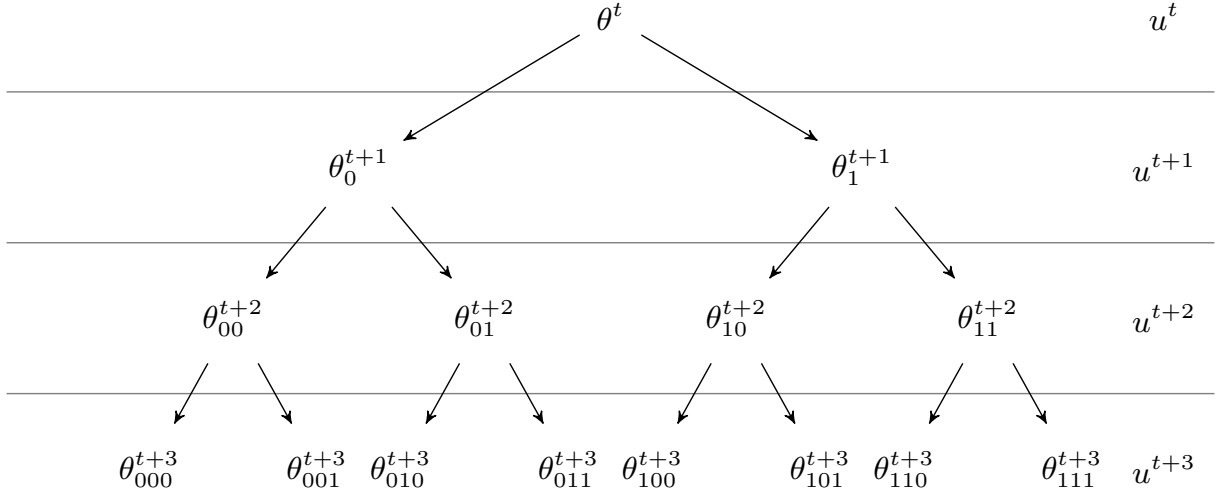

The second class of parallel MCMC algorithms uses parallelism through speculative execution to accelerate individual chains. This idea is called \emph{prefetching} in some of the literature and appears to have received only limited attention. To the best of our knowledge, prefetching has only been studied in the context of MH-style algorithms where, at each iteration, a single new proposal is drawn from a proposal distribution and stochastically accepted or rejected. The typical body of a MH implementation is a loop containing a single conditional statement and two associated branches, and so it is straightforward to view the possible execution paths as a binary tree, as illustrated in Figure~\ref{tree}. The vanilla version of prefetching speculatively evaluates all paths in this binary tree~\citep{brockwell-2006-prefetching}. The correct path will be exactly one of these, so with~$J$ cores, this approach achieves a speedup of~$\log_2 J$ with respect to single core execution, ignoring communication and bookkeeping overheads.

Na\"{i}ve prefetching can be improved by observing that the two branches are not taken with equal probability. On average, the reject branch tends to be more probable; the classic result for the optimal MH acceptance rate is 0.234~\citep{gelman-1997-accept}, so most prefetching scheduling policies have been built around the expectation of rejection. Let~${\alpha \le 0.5}$ be the expected probability of accepting a proposal. \citet{byrd-2008-SMP} introduced a procedure, called ``speculative moves,'' that speculatively evaluates only along the ``reject'' branch of the binary tree; in Figure~(\ref{tree}), this corresponds to the left-most branch. In each round of their algorithm, only the first~$k$ out of~${J-1}$ extra cores perform useful work, where $k$ is the number of rejected proposals before the first accepted proposal, starting from the root of the tree. The expected speedup is then:
\begin{align*}
1 + \E(k) < 1 + \sum_{k=0}^\infty k (1-\alpha)^k \alpha
< 1 + \frac{1-\alpha}{\alpha} = \frac{1}{\alpha}\,.
\end{align*}
Note that the first term on the left is due to the core at the root of the tree, which always performs useful computation in the prefetching scheme. For an acceptance rate of ${\alpha=0.23}$, this scheme yields a maximum expected speedup of about~$4.3$. It achieves an expected speedup of about~$4$ with~$16$ cores, and thus is more limited than the na\"{i}ve prefetching policy since it essentially cannot take advantage of additional cores.  \citet{byrd-2010-speculative} later considered the special case where the evaluation of the likelihood function occurs on two timescales, slow and fast.  They call this method ``speculative chains;'' it modifies ``speculative moves'' so that whenever the evaluation of the likelihood function is slow,  any available cores are used to speculatively evaluate the subsequent chain, assuming the slow step resulted in an accept.

Further extensions to the na\"{i}ve prefetching scheme allocate cores according to the optimal ``tree shape'' with respect to various assumptions about the probability of rejecting a proposal, \ie by greedily allocating cores to nodes that maximize the depth of speculative computation expected to be correct~\citep{strid-2010-prefetching}. Below, we summarize Strid's schemes and reference related ideas. Static prefetching assumes a fixed acceptance rate; versions of this were proposed
earlier in the context of
simulated annealing~\citep{witte-1991-SA}.
Dynamic prefetching estimates the acceptance probabilities, \eg at each level of the tree by drawing empirical MH samples (100,000 in the evaluation), or at each branch in the tree by computing $\min\{\beta, r\}$ where $\beta$ is a constant (${\beta = 1}$ in the evaluation) and $r$ is an estimate of the MH ratio based on a fast approximation to the target function. Alternatively, Strid proposes using the approximate target function to identify the single most likely path on which to perform speculative computation. Strid also combines prefetching with other sources of parallelism to obtain a multiplicative effect.  To the best of our knowledge, these prefetching methods have been developed for MH algorithms and evaluated on up to 64 cores, although usually many fewer.

\section{Predictive Prefetching}

We propose predictive prefetching, an improved scheduling approach that accelerates exact MCMC. Like Strid's dynamic prefetching procedure, we also exploit inexpensive but approximate target evaluations. Unlike existing prefetching methods, we combine this with the fact that the random stream used by a MCMC algorithm can be generated in advance and thus incorporated into the estimates of the acceptance probabilities at each branch in the binary tree.

\subsection{Mathematical Setup}

Consider a transition operator~${T(\theta\to\theta')}$ which has~$\pi(\theta)$ as its stationary distribution on state space~$\Theta$.  Simulation of such an operator typically proceeds using an ``external'' source of pseudo-random numbers that can, without loss of generality, be assumed to be drawn uniformly on the unit hypercube, denoted as~$\mcU$.  The transition operator is then a deterministic function from the product space of~$\mcU$ and~$\Theta$ back to~$\Theta$, \ie~${T:\Theta\times\mcU\to\Theta}$.  Most practical transition operators -- Metropolis--Hastings, slice sampling, {\it etc.} -- are actually compositions of two such functions, however.  The first function produces a countable set of candidate points in~$\Theta$, here denoted~${Q:\Theta\times\mcU_Q\to\mcP(\Theta)}$, where~$\mcP(\Theta)$ is the power set of~$\Theta$.  The second function~${R:\mcP(\Theta)\times\mcU_R\to\Theta}$ then chooses one of the candidates for the next state in the Markov chain.  Here we have used~$\mcU_Q$ and~$\mcU_R$ to indicate the disjoint subspaces of~$\mcU$ relevant to each part of the operator.  In this setup, the basic Metropolis--Hastings algorithm uses~$Q(\cdot)$ to produce a tuple of the current point and a proposed point, while
multiple-try MH~\citep{Liu2000}
and delayed-rejection MH~\citep{Tierney99, Green01}
create a larger set that includes the current point.  In the
exponential-shrinkage variant of slice sampling~\citep{Neal03},
the function~$Q(\cdot)$ produces an infinite sequence of candidates that converges to, but does not include, the current point.

This setup is a somewhat more elaborate treatment than usual, but this is intended to serve two purposes: 1)~make it clear that there is a separation between generating a set of possible candidates via~$Q(\cdot)$ and selecting among them with~$R(\cdot)$, and 2)~highlight that both of these functions are deterministic functions, given the pseudo-random variates.
Others have pointed out this ``deterministic given the randomness'' view,
and used it to construct alternative approaches to MCMC~\citep{propp-wilson-1996a, neal-2012a}.

We separately consider~$Q(\cdot)$ and~$R(\cdot)$ because it is generally the case that~$Q(\cdot)$ is inexpensive to evaluate and does not require computation of the target density~$\pi(\Theta)$, while~$R(\cdot)$ must compare the target density at the candidate locations and so represents the bulk of the computational burden.  Prefetching MCMC observes that, since~$Q(\cdot)$ is cheap and the pseudo-random variates can be produced in any order, the tree of possible future states of the Markov chain can be constructed before any of the~$R(\cdot)$ functions are evaluated, as in Figure~\ref{tree}.  The sequence of~$R(\cdot)$ evaluations simply chooses a path down this tree.  Parallelism can be achieved by speculatively choosing to evaluate~$R(\{\theta_i\}, u)$ for some part of the tree that has not yet been reached.  If this node in the tree is eventually reached, then we achieve a speedup.

For clarity, in the remainder of the paper we will focus on the straightforward random-walk Metropolis--Hastings operator.  In this special case,~$Q(\cdot)$ produces a tuple of the current point and a proposal, and the function~${R:\Theta\times\Theta\times(0,1)\to\Theta}$ takes these two points, along with a uniform random variate in~$(0,1)$, and selects one of the two inputs via:
\begin{align}
R(\theta,\theta',u) &= \begin{cases}
 \theta' & \text{if } u\frac{q(\theta'\given\theta)}{q(\theta\given\theta')} < \frac{\pi(\theta')}{\pi(\theta)}\\
 \theta & \text{otherwise}
 \end{cases}\,,
 \label{eqn:mh-threshold}
\end{align}
where~$q(\cdot\given\cdot)$ is the proposal density corresponding to~$Q(\cdot)$.  We write the acceptance ratio in this somewhat unusual fashion to highlight the fact that the left-hand side of the inequality does not require evaluation of the target density and is easy to precompute.

\subsection{Exploiting Predictions}

\newcommand{\Indicator}[1]{\iota_{#1}}
\newcommand{\Predictor}[1]{\psi_{#1}}
\newcommand{\PredEst}[2]{\psi_{#1}^{(#2)}}

A prefetching framework with $J$ cores uses one core to compute the immediate transition, and the others to precompute transitions for possible future iterations.
If each precomputation falls along the actual Markov chain, the framework will achieve the ideal linear speedup (evaluating $t$ iterations will take time proportional to $t/J$).
If some of them do not fall along the chain, the framework will fail to scale
perfectly with the available resources. For instance, a framework that evaluates transitions based on breadth-first search of the prefetching state tree (Figure~\ref{tree}) will achieve logarithmic speedup (time proportional to $t/\log_2 J$).
Good speedup thus depends on making good predictions of what path will be taken on the tree, which is in turn determined by our prediction of whether the threshold will be exceeded in Eq.~\ref{eqn:mh-threshold}. 

Let~$\rho$ denote a node on the tree,~$\theta_\rho$ indicate the current state at~$\rho$, and~$\theta'_\rho$ indicate the proposal.  We define
\begin{align}
r_\rho &= u_\rho \frac{q(\theta'_\rho \given \theta_\rho)}{q(\theta_\rho \given \theta'_\rho)}
\label{eqn:precomputed-threshold}
\end{align}
where~$u_\rho$ is the MH threshold variate for node~$\rho$.
The Markov chain's steps are determined by iterations of computing the indicator function ${\Indicator{\rho} = \mathbb{I}(r_\rho < \pi(\theta'_\rho)/\pi(\theta_\rho))}$, where a proposal is accepted iff ${\Indicator{\rho} = 1}$.
The quantities~$\theta_\rho$, $\theta'_\rho$, and~$r_\rho$ can be inexpensively computed at any time from the stream of pseudo-random numbers, without examining the expensive target~$\pi(\cdot)$. $r_\rho$ depends only on the depth (iteration) of~$\rho$.

The precomputation schedule should maximize expected speedup, which corresponds to the expected number of precomputations along the true path.
To maximize this quantity, the framework needs to anticipate which branches of the tree are likely to be taken. We associate with each node a predictor $\Predictor{\rho}$, where
\begin{align}
  \Predictor{\rho} & \approx \Pr\left( r_\rho < \frac{\pi(\theta'_\rho)}{\pi(\theta_\rho)}\right).
  \label{eqn:estimator}
\end{align}
This predictor may vary over time. When the target functions $\pi(\theta)$ and $\pi(\theta')$ are completely evaluated, we require that the predictor converges to the indicator $\Indicator{\rho}$.
Assuming that the predictions at each node are independent, then the probability that a node's computation is used is the product of its ancestors' predictors. Those nodes with maximum probability should be scheduled for precomputation. (The immediate transition will always be chosen: it has no ancestors and probability 1.) 

A predictor is always available -- for instance, one can use the recent acceptance probability -- but many problems can improve predictions using computation. To model this, we define a sequence of estimators
\begin{align}
  \PredEst{\rho}{m} &\approx \Predictor{\rho}, \quad m=0,1,2,\dots,N,
  \label{eqn:estimator-sequence}
\end{align}
where increasing $m$ implies increasing accuracy, and $\PredEst{\rho}{N} = \Indicator{\rho}$.
Workers move through this sequence until they perform the exact computation. The predictor sequence affects scheduling decisions: once it becomes sufficiently certain that a worker's branch will not be taken, that worker and its descendants should be reallocated to more promising branches.
Ultimately, every step that is actually taken on the Markov chain is computed to completion.  
The approach simulates from the true stationary distribution, not an approximation thereof. 
The estimators are used only in prefetching.

There are several schemes for producing this estimator sequence, and predictive prefetching applies to any Markov chain Monte Carlo problem for which approximations are available.
We focus on the important case where improved estimators are obtained by including more and more of the data in the posterior target distribution,
and in particular on Bayesian inference, a common and challenging task.

\subsection{Large-Scale Bayesian Inference}

In Bayesian inference with MCMC, the target density is a (possibly unnormalized) posterior distribution.  In most modeling problems, such as those corresponding to graphical models, the target density can be decomposed into a product of terms.  If the data are conditionally independent given the model parameters, there is a factor for each of the~$N$ data:
\begin{align}
\pi(\theta \given \x) &\propto \pi_0(\theta) \,\pi(\x \given \theta) 
                 = \pi_0(\theta) \prod_{n=1}^{N} \pi(x_n \given \theta)\,.
\end{align}
Here~$\pi_0(\theta)$ is a prior distribution and~$\pi(x_n\given\theta)$ is the likelihood term associated with the~$n$th datum.  The logarithm of the target distribution is a sum of terms
\begin{align}
\mcL(\theta) = \log \pi(\theta \given \x) &= \log \pi_0(\theta) + \log \pi(\x \given \theta) + c\nn \\
                 &= \log \pi_0(\theta) + \sum_{n=1}^{N} \log \pi(x_n \given \theta) + c\,, \nn
\end{align}
where~$c$ is an unknown constant that does not depend on~$\theta$ and can be ignored.
Our predictive prefetching algorithm uses this to form predictors~$\Predictor{\rho}$ as in Eq.~\ref{eqn:estimator}.  We can reframe~$\Predictor{\rho}$ using log probabilities as
\begin{align}
    \Predictor{\rho} &\approx \Pr\left(\log r_\rho < \mcL(\theta') - \mcL(\theta)\right),
\end{align}
where~$r_\rho$ is the precomputed random MH threshold of Eq.~\ref{eqn:precomputed-threshold}.  
One approach to forming this predictor is to use a normal model for each~$\mcL(\theta)$, as in \citet{Korattikara2014}.  However, rather than modeling~$\mcL(\theta)$ and~$\mcL(\theta')$ separately, we can achieve a better estimator with lower variance by considering them together.  Expanding each log likelihood, we get
\begin{align}
\mcL(\theta') - \mcL(\theta) &= \log\pi_0(\theta') - \log\pi_0(\theta) + \sum_{n=1}^N\Delta_n\\
\Delta_n &= \log\pi(x_n\given\theta') - \log\pi(x_n\given\theta)\,.
\end{align}
In Bayesian posterior sampling, the proposal $\theta'$ is usually a perturbation of $\theta$ and so we expect $\log \pi(x_n\given \theta')$ to be correlated with $\log \pi(x_n \given \theta)$.
In this case, the differences $\Delta_n$ occur on a smaller scale 
and have a smaller variance compared to the variance due to $\log \pi(x_n \given \theta)$ across data terms.

A concrete sequence of estimators is obtained by subsampling the data.
Let $\{\Delta_n\}_{n=1}^{m}$ be a subsample of size~${m<N}$, without replacement, from $\{\Delta_n\}_{n=1}^{N}$.  This subsample can be used to construct an unbiased estimate of~${\mcL(\theta') - \mcL(\theta)}$.
We model the terms of this subsample as i.i.d. from a normal distribution with bounded variance $\sigma^2$, leading to:
\begin{align}
  {\mcL(\theta') - \mcL(\theta)} &\sim \N(\hat{\mu}_m, \hat{\sigma}_m^2)\,.
\label{model-single}
\end{align}
The mean estimate $\hat\mu_m$ is empirically computable:
\begin{align}
  \hat{\mu}_m &= \log \pi_0(\theta') - \log\pi_0(\theta) + \frac{N}{m}\sum_{n=1}^m \Delta_n\,.
  \label{eqn:mu}
\end{align}
The error estimate $\hat\sigma_m$ may be derived from $s_m/\sqrt{m}$, where $s_m$ is the empirical standard deviation of the~$m$ subsampled~$\Delta_n$ terms. To obtain a confidence interval for the sum of~$N$ terms, we multiply this estimate by~$N$ and the finite population correction~$\sqrt{(N - m) / N}$, giving:
\begin{align}
\hat{\sigma}_m = s_m  \sqrt{\frac{N (N - m)}{m}}\,.
\label{eqn:sigma}
\end{align}
We can now form 
the predictor~$\PredEst{\rho}{m}$ by considering the tail probability for~$\log r_\rho$:
\begin{align}
\PredEst{\rho}{m} &= \int^\infty_{\log r_\rho}
\N(z\given \hat{\mu}_m, \hat{\sigma}_m^2)\,\mathrm{d}z\\
&= \frac{1}{2}\left[
1 + \erf\left( \frac{\log\hat{\mu}_m - \log r_\rho}{\sqrt{2}\hat{\sigma}_m}\right)
\right]\,.
\label{eqn:subset}
\end{align}

\subsection{System}

Our system uses these results as follows.
A master node manages the prefetching state tree and distributes a different node in the tree to each worker.
The worker given node~$\rho$ computes the corresponding proposal~$\theta'_\rho$ (which may consume values from the random sequence).
It asynchronously transmits the proposal, and the new point in the random sequence, back to the master.
It then starts evaluating the target function via progressively improved estimates, which it periodically reports back to the master.
Meanwhile, the master uses estimates of ${{\mcL}(\theta'_\rho) - \mcL(\theta_\rho)}$ values, the appropriate~$r_\rho$ constants, and an adaptive estimate of the current acceptance probability to calculate the predictor~$\PredEst{\rho}{m}$ for each node in the evaluation tree.
The master assigns workers to execute the target function only for those nodes most likely to be on the true path.
As estimates improve, some workers' proposals become less likely. Workers abandon unlikely proposals in favor of more likely ones. If the abandoned proposal becomes likely again, a worker will pick it up where the earlier worker left off.

In our implementation, the target posteriors~$\log\pi(\theta\given\x)$ and~$\log\pi(\theta'\given\x)$ are evaluated by separate workers.
Our normal model for the MH ratio based on a subsample of size $m$
depends on the empirical mean and standard deviation of
the differences~$\Delta_n$, but we use an approximation to avoid the extra
communication required to keep track of all these differences.
The worker for $\theta$ calculates
\begin{align}
  G_m(\theta) &= \log\pi_0(\theta) + \frac{N}{m}\sum_{n=1}^m \log\pi(x_n\given\theta)
  \label{eqn:implmu}
\end{align}
rather than the difference mean $\hat\mu_m$ from Eq.~\ref{eqn:mu}. Given these values, the master can precisely compute~${\hat{\mu}_m = G_m(\theta') - G_m(\theta)}$,
but the empirical standard deviation of differences, $s_m$ in Eq.~\ref{eqn:sigma}, must be estimated. We set
\begin{align}
s_m = \sqrt{S_m(\theta)^2 + S_m(\theta')^2 - 2 \tilde{c} S_m(\theta) S_m(\theta')}\,,
\end{align}
where $S_m(\theta)$ denotes the empirical standard deviation of
the $m$ $\log \pi(x_n \given \theta)$ terms, and $\tilde{c}$ 
approximates the correlation between $\log \pi(x_n \given \theta)$
and $\log \pi(x_n \given \theta')$.
We empirically observe this correlation to be very high;
in all experiments we set $\tilde{c} = 0.9999$.
Note that this approximation only affects the quality of our speculative
predictions; it does not affect the actual decision to accept or reject the
proposal $\theta'$.

Our implementation requires at least two cores, one master and one worker (although when there is only one worker the master is basically irrelevant).

\begin{table}[t]
\centering
\begin{tabular}{@{}r  r@{~~}r@{\qquad}r@{~~}r@{\qquad}r@{~~}r@{}}
     & \multicolumn{2}{c@{\qquad}}{Burn-in}
     & \multicolumn{2}{c@{\qquad}}{}
     & \multicolumn{2}{c@{}}{}\\
$J$ & \multicolumn{2}{c@{\qquad}}{$i_1=9575$}
     & \multicolumn{2}{c@{\qquad}}{$i_2=24000$}
     & \multicolumn{2}{c@{}}{$i_3=50000$}\\
     \noalign{\vskip1pt}%
     \hline
     \noalign{\vskip2pt}%
     1 & 16674 & --- & 41978 & --- & 87500 & ---\\ 
     16 & 2730 & $6.1\times$ & 8678 & $4.3\times$ & 20318 & $4.3\times$ \\ 
     32 & 1731 & $9.6\times$ & 7539 & $5.6\times$ & 19046 & $4.6\times$ \\ 
     64 & 989 & $16.8\times$ & 5894 & $7.1\times$ & 15146 & $5.8\times$ \\
   \end{tabular}
   \caption{Cumulative time (in seconds) and speedup for evaluating the Gaussian mixture model with different numbers of workers $J$.}
\label{tab:gauss}
\end{table}

\section{Experiments}

Our evaluation focuses on MH for large-scale Bayesian inference using the approximations described above
(though our framework can use any approximation scheme for the target distribution).
Our implementation is written in C++ and Python,
and uses MPI for communication between the master and worker cores.
We evaluate our implementation on up to 64 cores in a multicore
cluster environment in which  machines are connected by 10GB ethernet
and each machine has 32 cores (four 8-core Intel Xeon E7-8837 processors).
We report speedups relative to serial computation with one worker.

\begin{figure}[t!]
\begin{center}
\includegraphics[width=0.49\textwidth]{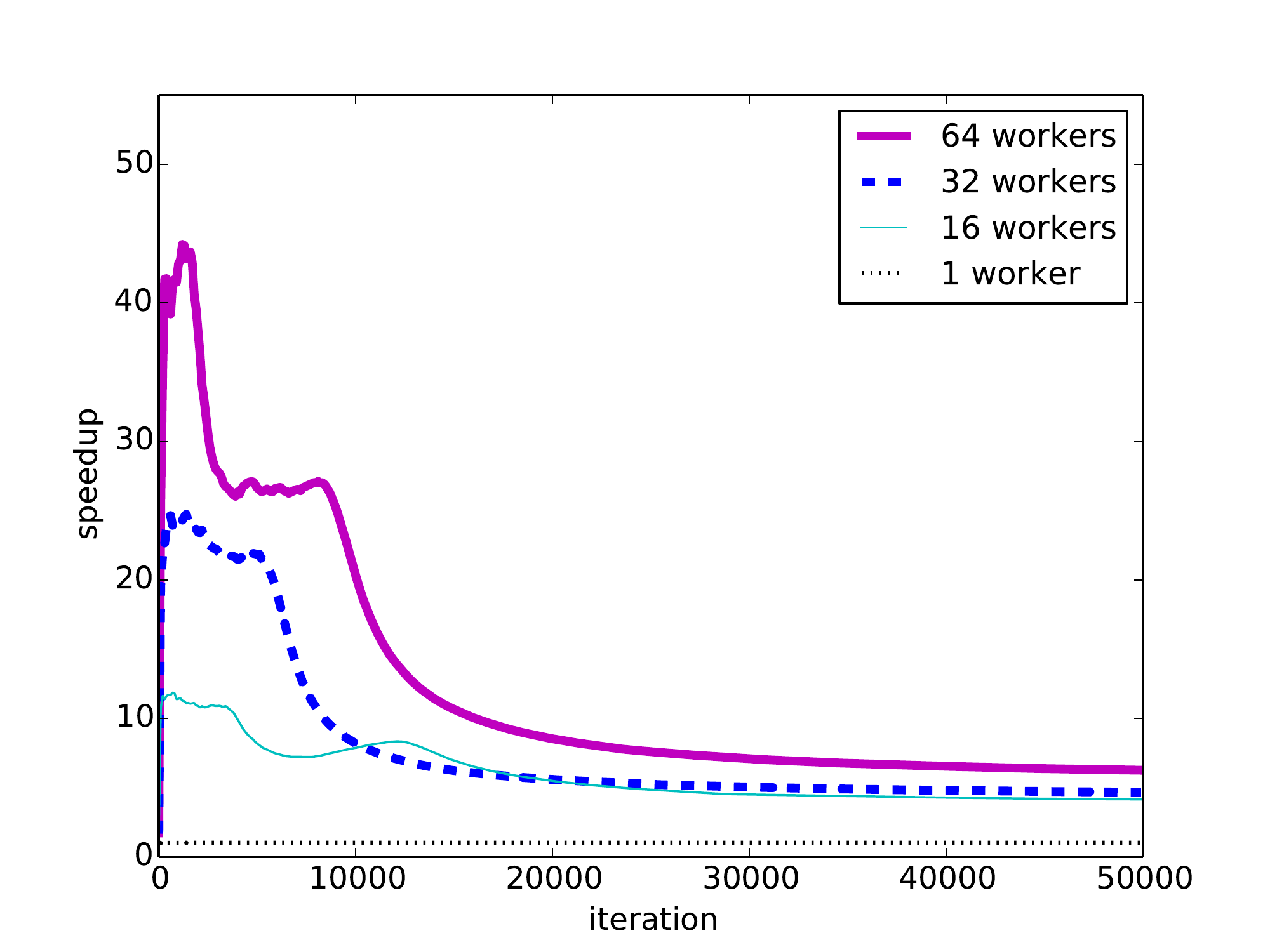}
\end{center}
\caption{Cumulative speedup relative to our baseline,
as a function of the number of MH iterations, for the mixture of Gaussians problem.
The different curves correspond to different numbers of workers.}
\label{fig:mixture}
\end{figure}

We evaluate our system on both synthetic and real Bayesian inference problems.
Our first target distribution is the posterior density of the eight-component
mixture of eight-dimensional Gaussians used by~\citet{nishihara-2012-gess},
where the likelihood involves $10^6$ samples drawn from this model.
Our second target distribution is the posterior density of
a Bayesian Lasso regression~\citep{Park08} that models molecular photovoltaic activity.
The likelihood involves a dataset of ${1.8 \times 10^6}$ molecules
described by 56-dimensional real-valued features;
each response is real-valued and corresponds to a lengthy density functional theory calculation \citep{cep}.

In all of our experiments, we use a spherical Gaussian for the proposal distribution.
A simple adaptive scheme sets the
scale of this distribution, improving convergence relative to standard MH.
Our approach falls under the provably
convergent adaptive algorithms
studied by~\citet{Andrieu06}
and was easily incorporated into our framework.

We expect predictive prefetching to perform best when the densities at a proposal and corresponding current point are significantly different, which is common in the initial burn-in phase of chain evaluation.
In this phase, early estimates based on small subsamples effectively predict whether the proposal is accepted or rejected.
When the density at the proposal is very close to that at the current point -- for example, as the proposal distribution approaches the target distribution -- the outcome is inherently difficult to predict; early estimates will be uncertain or even wrong.
Incorrect estimates could destroy speedup (no precomputations would be useful). We hope to do better than this worst case, and to at least achieve logarithmic speedup.

In our experiments, we divide the evaluation of the target
function into 100 batches. Thus, for the mixture problem, each subsample contains $10^4$ data items.

\begin{table}[t]
\centering
\begin{tabular}{@{}l r r r r@{}}
& & standard & & \\
& mean & deviation & min & max \\
\noalign{\vskip1pt}
\hline
\noalign{\vskip2pt}
$n_\text{eff}$ & 3405 & 7253 & 50 & 26000 \\
$\hat{R}$ & 1.005 &  0.006 & 1.000 & 1.020 \\
\end{tabular}
\caption{Convergence statistics after burn-in (over iterations $i_2$--$i_3$) for the Gaussian mixture model,
computed over the 64 dimensions of the model.}
\label{tab:convergence}
\end{table}

Table~\ref{tab:gauss} shows the results for the Gaussian mixture model.
We run the model with the same initial conditions and pseudorandom sequences with varying numbers of worker threads. All experiments produce identical chains.
We evaluate the cumulative time and speedup obtained at three different iteration counts.
The first, ${i_1 = 9575}$ iterations, is burn-in. After $i_1$ iterations, all dimensions of samples achieve the Gelman-Rubin statistic ${\hat R < 1.05}$, computed using two independent chains, where the first $i_1/2$ samples have been discarded \citep{gelman1992inference}.
We then run the model further to $i_3$ iterations. Iterations ${i_2 = 24000}$ through ${i_3 = 50000}$ are used to compute an effective number of samples $n_\text{eff}$. (Table~\ref{tab:convergence} shows convergence statistics after $i_3$ iterations.)
The results are as we hoped. The initial burn-in phase obtains better-than-logarithmic speedup (though not perfect linear speedup).  With 64 workers, the chain achieves burn-in 16.8$\times$ faster than with one worker.
After burn-in, efficiency drops as expected, but we still achieve logarithmic speedup (rather than sub-logarithmic). At $50000$ iterations, speedup for each number of workers $J$ rounds to $\log_2 J$.

\begin{figure}[t!]
\begin{center}
\includegraphics[width=0.49\textwidth]{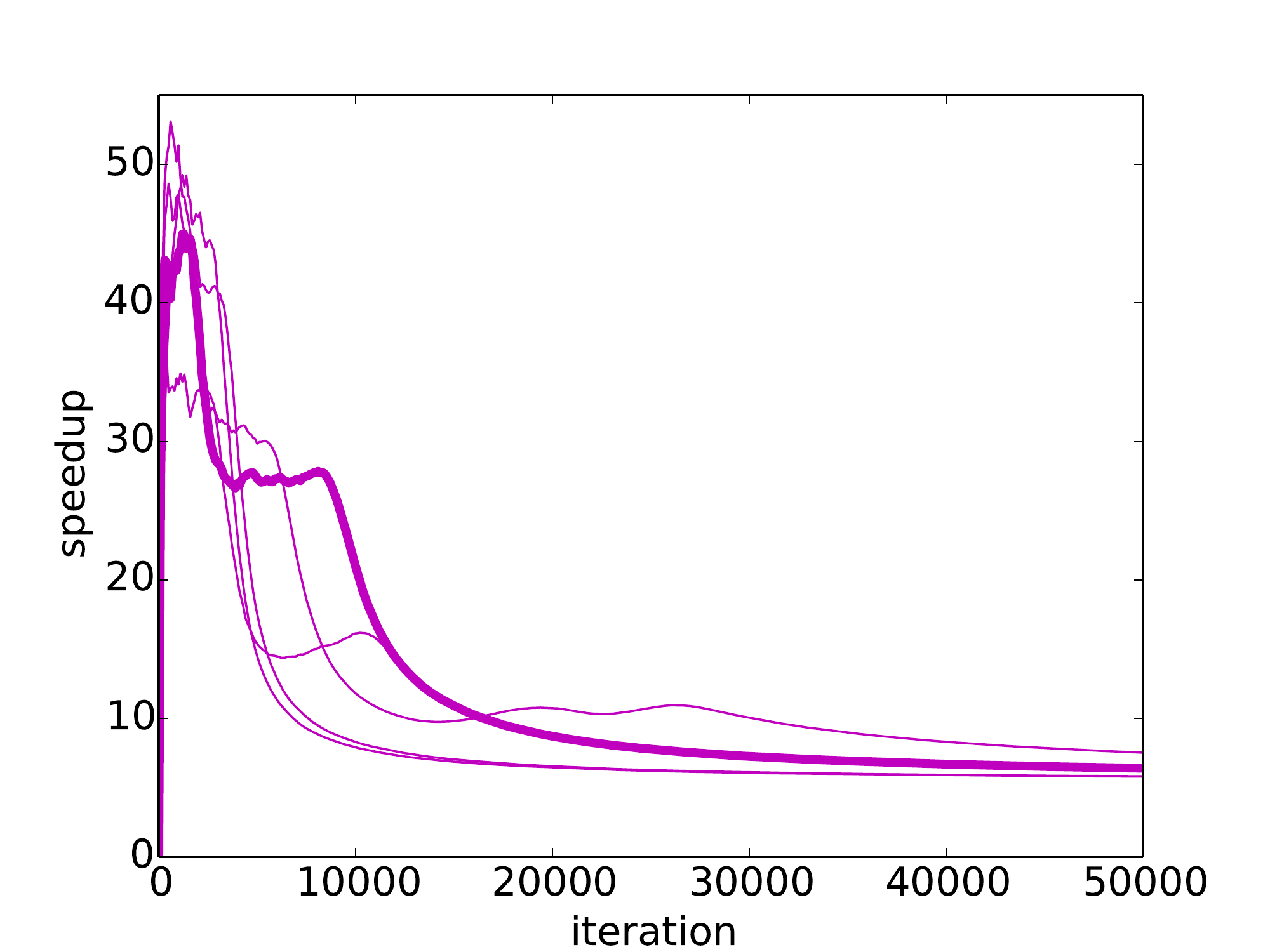}
\end{center}
\caption{Cumulative speedup relative to our baseline,
as a function of the number of MH iterations, for the mixture of Gaussians problem.
The different curves correspond to different initial conditions;
all curves are for 64 workers.}
\label{fig:initcond}
\end{figure}

Figure~\ref{fig:mixture} explains these results by graphing cumulative speedup over the whole range of iterations.
The initial speedup is close to linear -- we briefly achieve more than $40\times$ speedup at $J = 64$ workers.
As burn-in proceeds, cumulative speedup falls off to logarithmic in $J$.
Figure~\ref{fig:initcond} shows cumulative speedup for the Gaussian mixture model with several different initial conditions.
We see a range of variation due to differences in the adaptive scheme during burn-in.
The overall pattern is stable, however: good speedup during burn-in followed by logarithmic speedup later.
Also note that speedup does not necessarily decrease steadily, or even monotonically. At some initial conditions, the chain enters an easier-to-predict region before truly burning in; while in such a region, speedup is maintained. Our system takes advantage of these regions effectively.

\begin{figure}[t!]
\begin{center}
\includegraphics[width=0.49\textwidth]{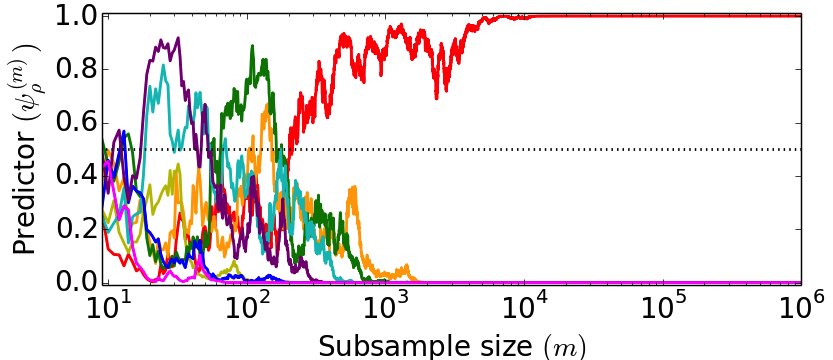}
\textbf{A. Burn-in}
\vskip.5\baselineskip
\includegraphics[width=0.49\textwidth]{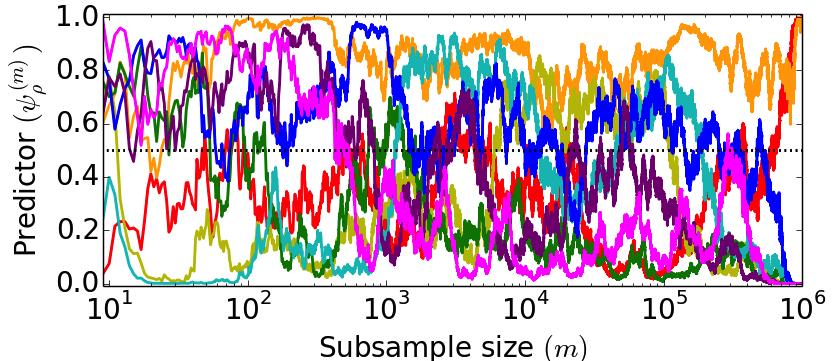}
\textbf{B. Convergence}
\end{center}
\caption{Example predictor trajectories for the mixture of Gaussians problem. We show the predictor $\PredEst{\rho}{m}$ as a function of subsample size $m$. Different colors indicate different proposals. Burn-in is much easier to predict than convergence.}
\label{fig:predictor}
\end{figure}

Figure~\ref{fig:predictor} shows how our predictors behave both during and after burn-in.
During burn-in, estimates are effective, and the predictor converges quite quickly to the correct indicator.
After burn-in, the new proposal's target density is close to the old proposal's, and the estimates are similarly hard to distinguish. Sometimes the random variate $r_\rho$ is small enough for the predictor to converge quickly to 1; more often, the predictor varies widely over time, and does not converge to $0$ or $1$ until almost all data are evaluated. This behavior makes logarithmic speedup a best case. Luckily, the predictor is more typically uncertain (with an intermediate value) than wrong (with an extreme value that eventually flips to the opposite value): incorrect predictors could lead to sublogarithmic speedup.

\begin{figure*}[t]
\centering%
\subfloat[]{%
\includegraphics[width=0.32\textwidth]{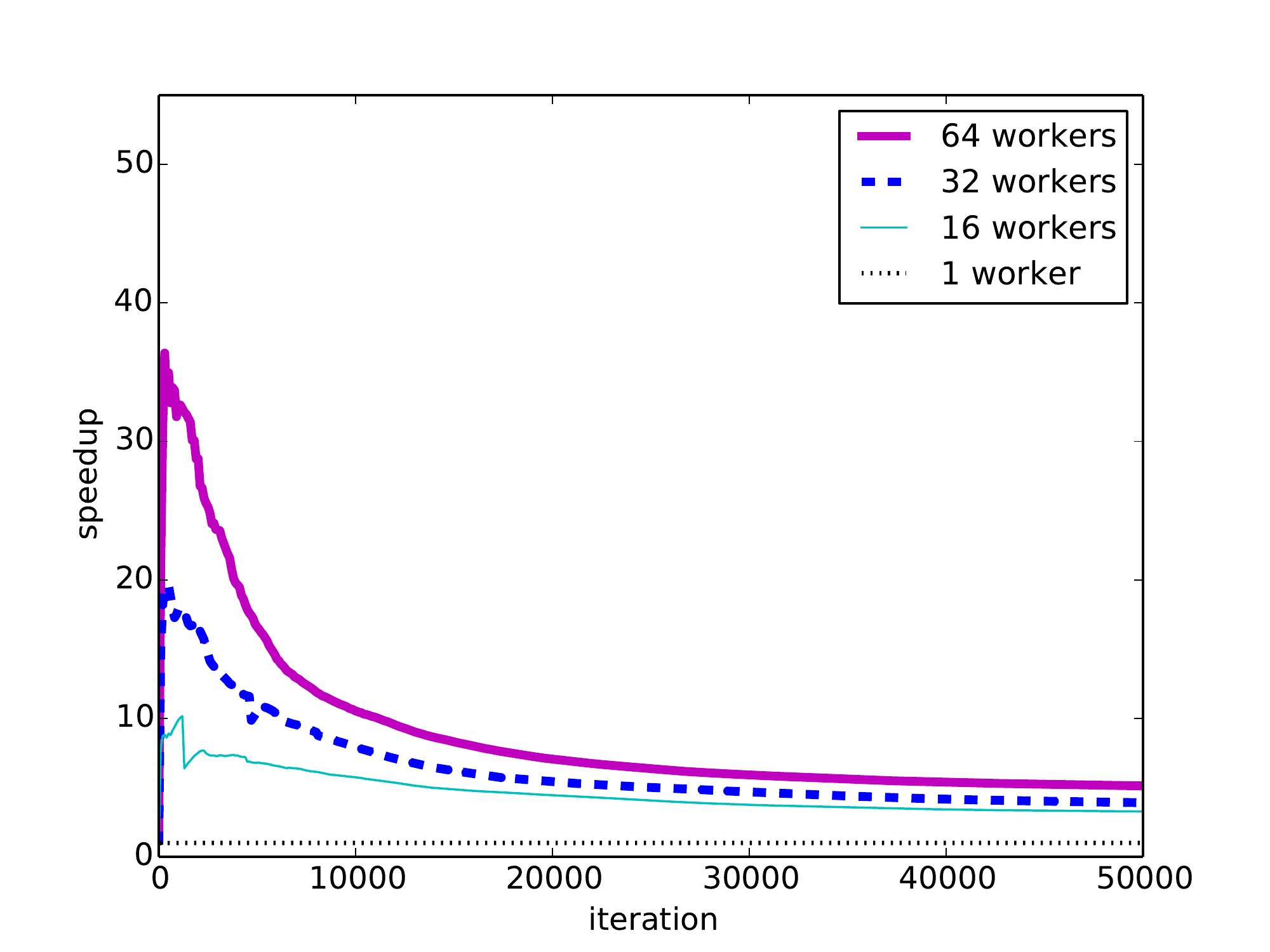}}~
\subfloat[]{%
\includegraphics[width=0.32\textwidth]{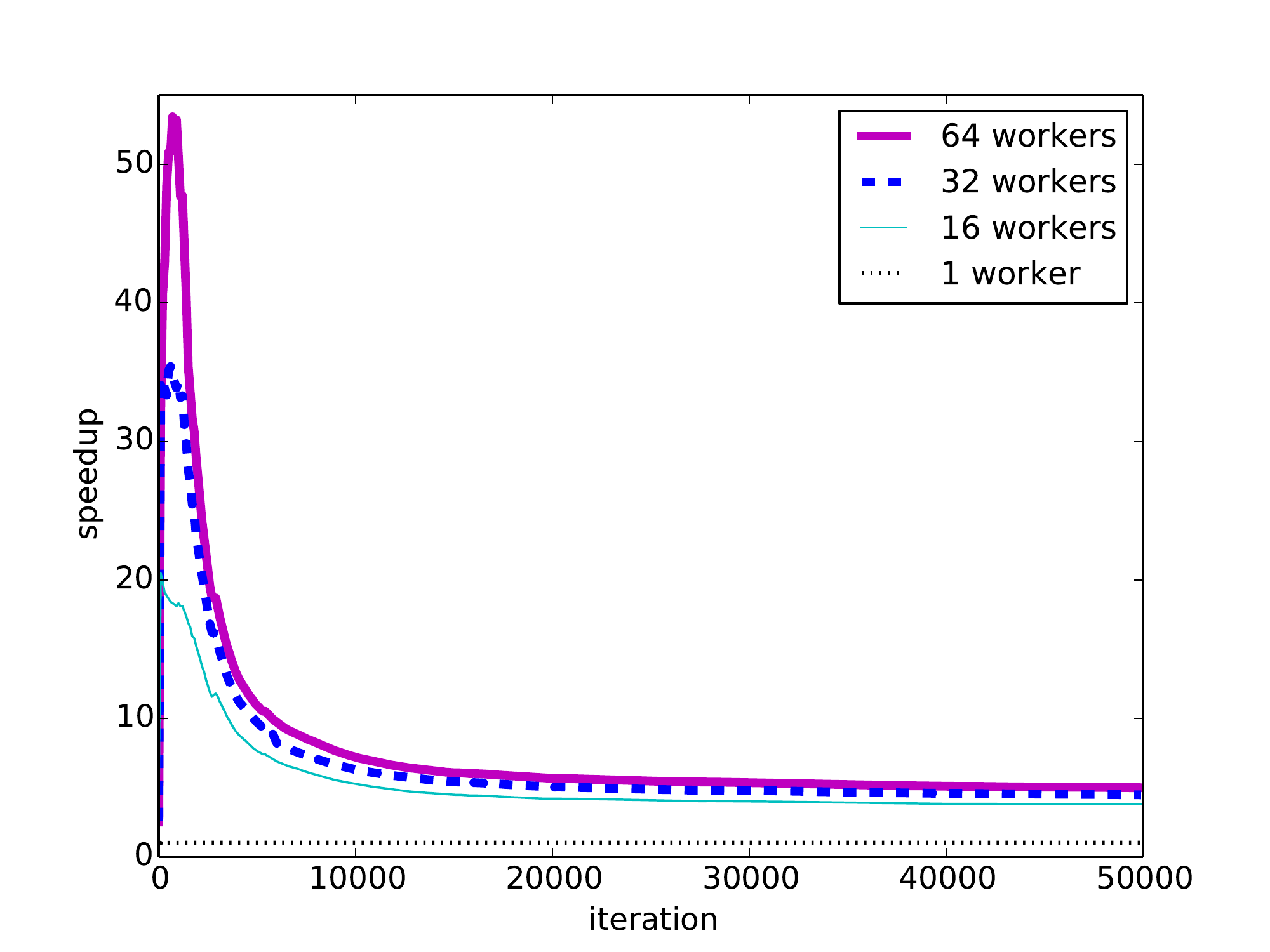}}~
\subfloat[]{%
\includegraphics[width=0.32\textwidth]{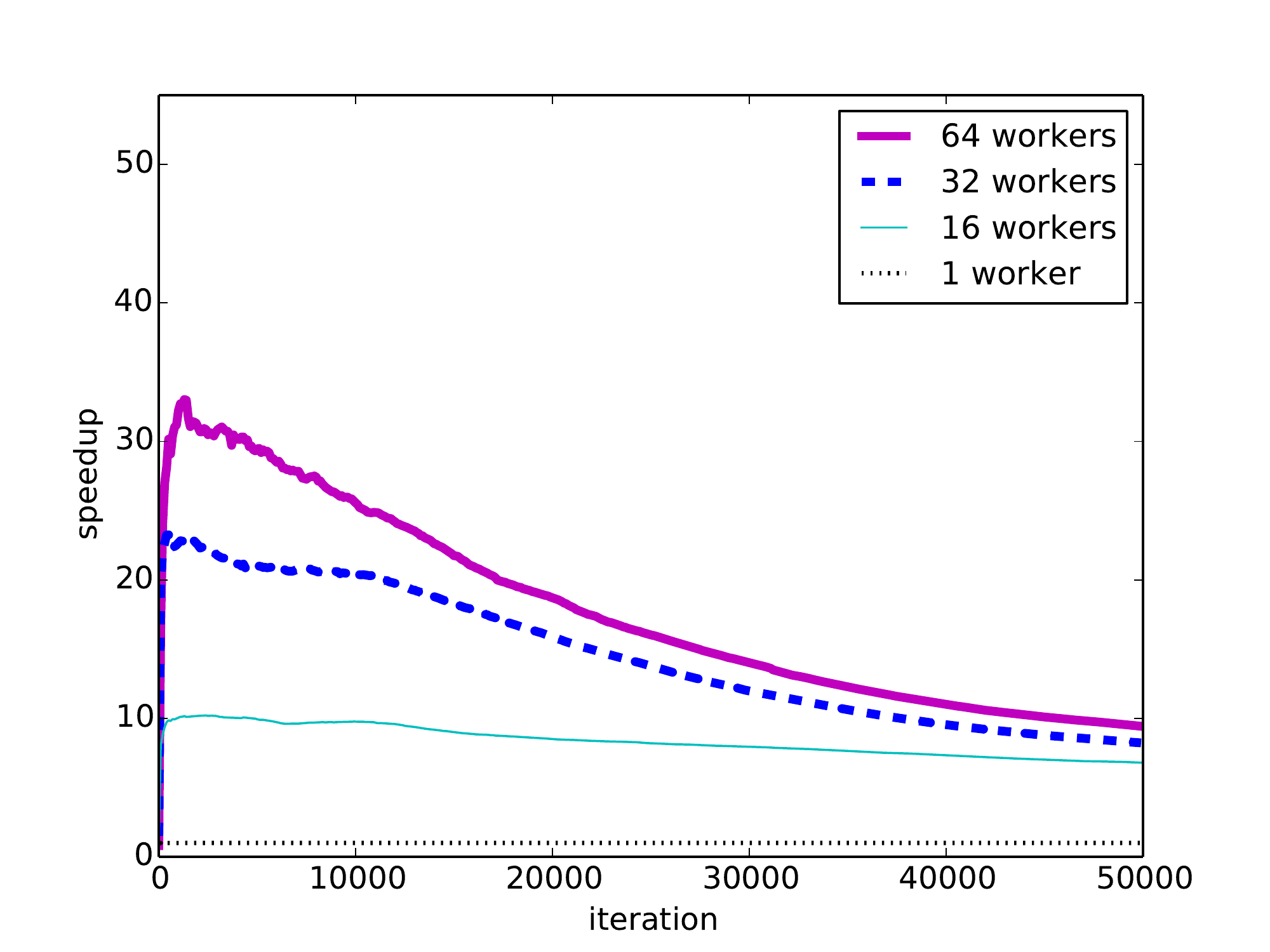}}
\caption{Cumulative speedup relative to our baseline,
as a function of the number of MH iterations, for the Bayesian Lasso problem.
The different curves correspond to different numbers of workers.
The different figures are for different initial conditions.}
\label{fig:lasso}
\end{figure*}

Figure~\ref{fig:lasso} shows that good speedups are achievable for real problems.
The speedup distribution for the Bayesian Lasso problem for molecular photovoltaic activity appears similar to that of the mixture of Gaussians.
There are differences, however: Lasso evaluation did not converge by 50000 iterations according to standard convergence statistics. On several initial conditions, the chain started taking small steps, and therefore dropped to logarithmic speedup, before achieving convergence.
Overall performance might be improved by detecting this case and switching some speculative resources over to other initial conditions, an idea we leave for future work.

\section{Conclusions}

We presented parallel predictive prefetching, a general framework for accelerating many widely used MCMC algorithms that are inherently serial and often slow to converge.
Our approach applies to MCMC algorithms whose transition operator can be decomposed into two functions: one that produces a countable set of candidate proposal states and a second that chooses the next state from among these.
Predictive prefetching uses parallel cores and speculative computation to exploit the common setting in which (1) generating proposals is computationally fast compared to the evaluation required to choose from among them and (2) this latter evaluation can be approximated quickly.
Our first focus has been on the MH algorithm, in which predictive prefetching exploits a sequence of increasingly accurate predictors for the decision to accept or reject a proposed state.
Our second focus has been on large-scale Bayesian inference, for which we identified an effective predictive model that estimates the likelihood from a subset of data.
The key insight is that we model the uncertainty of these predictions with respect to the difference between the likelihood of each datum evaluated at the proposal and current state.  
As these evaluations are highly correlated, the variance of the differences is much smaller than the variance of the states evaluated separately, leading to significantly higher confidence in our predictions.
This allows us to justify more aggressive use of parallel resources, leading to greater speedup with respect to serial execution or more na\"{i}ve prefetching schemes.

The best speedup that is realistically achievable for this problem is sublinear in the number of cores but better than logarithmic, and our results achieve this.
Our approach generalizes both to schemes that learn an approximation to the target density and to other MCMC algorithms with more complex structure, such as slice sampling and more sophisticated adaptive techniques.

\subsubsection*{Acknowledgments}
We thank M.P. Brenner, E.D. Cubuk, V. Kanade, Z. Liu, D. Maclaurin, A.C. Miller and R. Nishihara for helpful discussions.  This work was partially funded by DARPA Young Faculty Award N66001-12-1-4219, the
National Institutes of Health under Award Number 1R01LM010213-01 and Google.

\bibliography{refs}
\bibliographystyle{abbrvnat}

\end{document}